\documentclass[lettersize,journal]{IEEEtran}
\usepackage{amsmath,amsfonts}
\usepackage{algorithmicx}
\usepackage[ruled, linesnumbered]{algorithm2e}
\usepackage{array}
\usepackage[caption=false,font=normalsize,labelfont=sf,textfont=sf]{subfig}
\usepackage{textcomp}
\usepackage{stfloats}
\usepackage{url}
\usepackage{verbatim}
\usepackage{graphicx}
\usepackage{cite}
\usepackage{enumerate}
\usepackage{xspace}
\usepackage{hyperref}
\usepackage{cleveref}
\usepackage{multirow}
\usepackage{booktabs}
\begin{document}
\title{GPT-NAS: Evolutionary Neural Architecture Search with the Generative Pre-Trained Model}
\author{\IEEEauthorblockN{Caiyang Yu, Xianggen Liu, Yifan Wang, Yun Liu, Wentao Feng, Deng Xiong, \IEEEmembership{Member,~IEEE}, Chenwei Tang\IEEEauthorrefmark{1}, \IEEEmembership{Member,~IEEE} and Jiancheng Lv\IEEEauthorrefmark{1},~\IEEEmembership{Senior Member,~IEEE}}
\thanks{C. Yu, X. Liu, Y. Wang, Y. Liu, W. Feng, C. Tang and J. Lv are with the College of Computer Science, Sichuan University, Chengdu 610065, PR China (e-mail: yucy324@gmail.com, \{wangyifan5217, yliu\}@stu.scu.edu.cn, \{liuxianggen, Wtfeng2021, tangchenwei, lvjiancheng\}@scu.edu.cn). D. Xiong is with the Department of Mechanical Engineering, Stevens Institute of Technology, Hoboken, NJ 07030, USA. (e-mail: dxiong@stevens.edu.)}
\thanks{Corresponding author: Chenwei Tang and Jiancheng Lv}}

\markboth{Journal of \LaTeX\ Class Files,~Vol.~14, No.~8, August~2021}%
{How to Use the IEEEtran \LaTeX \ Templates}

\maketitle

\begin{abstract}
Neural Architecture Search (NAS) has emerged as one of the effective methods to design the optimal neural network architecture automatically. Although neural architectures have achieved human-level performances in several tasks, few of them are obtained from the NAS method. The main reason is the huge search space of neural architectures, making NAS algorithms inefficient. This work presents a novel architecture search algorithm, called GPT-NAS, that optimizes neural architectures by Generative Pre-Trained (GPT) model with an evolutionary algorithm (EA) as the search strategy. In  GPT-NAS, we assume that a generative model pre-trained on a large-scale corpus could learn the fundamental law of building neural architectures. Therefore, GPT-NAS leverages the GPT model to propose reasonable architecture components given the basic one and then utilizes EAs to search for the optimal solution. Such an approach can largely reduce the search space by introducing prior knowledge in the search process. Extensive experimental results show that our GPT-NAS method significantly outperforms seven manually designed neural architectures and thirteen architectures provided by competing NAS methods. In addition, our experiments also indicate that the proposed algorithm improves the performance of fine-tuned neural architectures by up to about 12\% compared to those without GPT, further demonstrating its effectiveness in searching neural architectures. 
\end{abstract}


\section{Introduction}
\label{Introduction}
\IEEEPARstart{I}{n} recent years, deep neural networks have shown impressive fitting power in various tasks, ranging across computer vision \cite{he2016deep, sun2019evolving}, natural language processing \cite{blanchard2022automating, zhang2023lifelong} and bioinformatics \cite{min2017deep}. In deep learning (DL), it is widely accepted that neural networks with individual architectures present different inductive biases. Although multiple advanced architectures have been designed, the intrinsic principles of the architectural building remain unclear. As a result, researchers usually consume large times to manually seek for the  neural networks that are suitable to the given tasks.

To accelerate the designing process and improve the quality of the neural architectures, neural architecture search (NAS) \cite{ren2021comprehensive} has emerged as one of the effective methods to design the optimal neural network architecture automatically. The main advantage of NAS lies in its ability to automate the tedious and time-consuming process of designing neural architectures. Additionally, NAS can improve the quality of neural architectures by leveraging search strategies to find architectures that achieve better performance than human-designed architectures. Currently, based on different optimization techniques, the mainstream NAS search strategies include reinforcement learning (RL) \cite{zoph2016neural, baker2016designing}, evolutionary algorithm (EA) \cite{zhang2021adaptive, sun2021novel, xue2023neural, peng2022pre}, and gradient optimization (GO) \cite{liu2018darts, pham2018efficient}. Algorithms such as NAS-RL \cite{zoph2016neural}, NASNet \cite{zoph2018learning}, MetaQNN \cite{baker2016designing} and Block-QNN-S \cite{zhong2018practical} all belong to the first category. For different RL methods, the key lies in how to design the agent's policy and the corresponding optimization process \cite{elsken2019neural}. For example, Zoph \textit{et al.} \cite{zoph2016neural} use the RNN policy to select the basic information and form the neural architecture, while the Proximal policy is used for optimization in subsequent work \cite{zoph2018learning}. In addition, Q-learning is used in \cite{baker2016designing, zhong2018practical} to train the policy. Secondly, the EA-based NAS (EA-NAS) searches for the optimal architecture mainly by the properties of the algorithm. For example, in \cite{xie2017genetic, sun2020automatically}, the genetic algorithm (GA) is used as the optimization strategy to complete the algorithm search process, while in \cite{suganuma2017genetic}, the genetic programming strategy is adopted. Finally, GO-based NAS is a category of algorithms that do not rely on any strategy. It mainly implements search in a continuous search space, such as \cite{liu2018darts}.

While neural architectures have achieved human-level performances in several tasks, only a few of them have been obtained from the NAS method. The main challenge with NAS is that its effectiveness is often hindered by the vast search space of possible architectures. The sheer number of possible architectures quickly becomes unmanageable and poses a significant obstacle to the search process. Consequently, the search strategy used in NAS can become less effective and make it exceedingly difficult to find optimal architectures. Additionally, the search for an optimal architecture is further complicated when factors such as computational efficiency, model size and accuracy need to be considered simultaneously.


As a result, several improvement efforts have been generated to adjust the search space in the search process. In \cite{ci2021evolving}, a Neural Search-space Evolution (NSE) scheme is proposed for large search spaces. Instead of starting directly from a huge search space, NSE obtains a subset from the search space and then searches it to obtain an optimized space. The quality of the optimized search space is improved by continuously evolving the candidate operations in the subset. The step is repeated until the entire large search space is traversed. Xue \textit{et al.} \cite{xue2022automated} found that the search space and the search strategy are coupled with each other, and the two can search for better-form algorithms if they can reasonably cooperate. As a result, a method to automatically select the search strategy and the search space is proposed to solve both problems simultaneously.

Although the above methods propose new solutions for the large search space, they still carry risks. For example, in \cite{ci2021evolving}, the authors train the supernet in a subset of the search space and will drop some candidate operations based on the fitness, but there is no guarantee that the dropped candidate solutions are invalid for the whole search space, so there will be false drops. In addition, dynamically adjusting or reducing the size of the search space during the search process would result in missed opportunities to discover novel and effective neural architecture and it would also complicate the implementation of the algorithm.

To this end, we propose a NAS algorithm based on Generative Pre-trained Transformer (GPT-NAS), which is an innovative solution for the large search space. Unlike traditional approaches that focus solely on the search space or search strategy, our proposed GPT-NAS algorithm leverages the power of GPT \cite{radford2018improving} models to introduce a priori knowledge into the algorithm. This knowledge serves as a valuable guide for the search process, reducing uncertainty and effectively shrinking the search space. 
On the other hand, the GPT model is widely recognized for their extraordinary ability to learn complex patterns and generate high-quality sequences, especially with the advent of Large Language Models (LLMs) \cite{zhao2023survey}. As exemplified by ChatGPT, these LLMs have shown excellent performance in a wide range of tasks and domains, from language modelling to image recognition. This also provides a strong backbone to our work.

In the work, we design a GPT model with neural architectures as input data and effectively combine the GPT model with EA-NAS. 
The GPT-NAS algorithm can be divided into three procedures: neural architecture encoding, pre-training and fine-tuning the GPT model, and network architecture search.
Specifically, firstly, the GPT-NAS algorithm employs an encoding strategy to map the neural architecture to a vector representation, which is then used as input to the GPT model. Secondly, the GPT model is pre-trained on a large amount of neural architecture data to learn the patterns and relationships between different architectural components and operations. Then it is fine-tuned for specific tasks. Finally, the fine-tuned GPT model is used for network architecture search, in which we use GA as the search strategy to achieve the search of the optimal architecture.
With the GPT-NAS algorithm, poor-quality neural architectures can be quickly dropped, thus reducing the search space and increasing the efficiency of the NAS algorithm. In addition, for GPT-NAS, an acceleration strategy has been proposed in order to reduce the cost of running the algorithm.


We evaluated our model on the CIFAR-10, CIFAR-100 \cite{krizhevsky2009learning}, and ImageNet-1K \cite{ILSVRC15}. Experimental results show that GPT-NAS is able to demonstrate excellent performance on three datasets. A summary of specific contributions of this article is presented as follows.

\begin{enumerate}[(1)]
\item We propose a novel NAS algorithm, called GPT-NAS, that uses the GPT model to guide the search for neural architectures in the algorithm. To the best of our knowledge, we are the first to propose the introduction of GPT into a NAS algorithm.

\item We propose to use a large number of neural architectures as input to train the GPT model, enabling it to learn the properties and characteristics of different network designs. This can help improve the performance of the GPT model in tasks related to architecture search for neural networks.

\item We propose a specific acceleration strategy for GPT-NAS that can significantly reduce the time required for searching the best algorithm. This strategy can help improve the efficiency of the GPT-NAS algorithm and enable faster neural architecture search.

\item Extensive experiments have proven that GPT-NAS demonstrates state-of-the-art experimental results on three datasets. In addition, we also demonstrate that the proposed acceleration strategy can effectively reduce the time cost.


\end{enumerate}

The remainder of the article is presented below. The literature review is discussed in Section \ref{Literature review}. Section \ref{Proposed Method} documents the proposed method, including the encoding strategy, the design of the GPT model, and acceleration strategies. Section \ref{Experimental Design} describes the experimental design. Section \ref{Experimental Results} discusses the experimental results. The conclusion and future work are present in Section \ref{Conclusion and Future work}.

\section{Literature Review}
\label{Literature review}

In this section, the GPT model \cite{radford2018improving}, which is the essential technique of this study, will be introduced in the background. In addition, we will also review the specific knowledge of NAS in the related work to help the reader understand it better.

\subsection{Background}
\subsubsection{Generative Pre-Training} The GPT is a state-of-the-art (SOTA) language model that utilizes unsupervised learning to pre-train on a vast amount of text data. By leveraging this massive amount of data, GPT can extract a wide range of common features and patterns that exist within the language, such as syntax, grammar, and semantics. This pre-training approach makes GPT a highly versatile model that can be fine-tuned for a wide variety of natural language processing tasks, such as language translation \cite{hendy2023good}, sentiment analysis \cite{mathew2020review}, and question answering \cite{bongini2023gpt}, to name a few. Because GPT has already learned many of the fundamental features of language, it can take on specific tasks with a lighter learning burden than models trained from scratch, enabling it to produce higher-quality results in less time. In terms of structure, the GPT model uses the decoder of Transformer \cite{vaswani2017attention} and makes some changes to it. The original decoder has two Multi-Head Attention, while the changed decoder has only one Mask Multi self-attention. The purpose is that the model employs the preceding part of the text to predict the next, and the Mask Multi self-attention used in the decoder can mask the following data. For example, giving the sentence ``\textit{I am a student.}", when GPT predicts the word ``\textit{a}", only ``\textit{I am}" can be entered as input, and the remaining words ``\textit{a student.}" need to be masked. Hence, it fits the current application scenario in this paper. 

The training process of the GPT model can be divided into two phases, unsupervised pre-training in the first phase and supervised fine-tuning in the second phase. In the pre-training phase, the sentences are embedded and fed into the Transformer decoder to learn a language model, while in the fine-tuning phase, the pre-training parameters are tuned to the specific task. The overall purpose of the model is to learn a general representation method that can be adapted to different kinds of tasks with only minor modifications.




\subsection{Related Work}

In this paper, we propose a new generative idea for neural network architectures, so we will present the related work in this area in detail.
\subsubsection{Neural Architecture Generation}

NAS algorithms vary in the form of neural architecture generation, ranging from those composed based on layer or block structures to those obtained based on smaller base units, such as convolutional kernels, strides, and other hyperparameters. Among these, NAS-RL \cite{zoph2016neural} is one of the most classical algorithms, and at first glance, the idea in this study is very similar to it.

In NAS-RL, the author proposed to use a recurrent neural network (RNN) as a controller to generate the neural architecture (child network), while the controller is trained by RL in order to maximize the performance of the generated neural architecture. Specifically, one or more layers in the RNN represent the parameter information of each layer of the child network, which is viewed as actions in RL. In addition, the accuracy of the child network is considered a reward signal and is used to compute the policy gradient to update the controller. To increase the complexity of the child network, skip connections and other layer types are introduced to make it more competitive. Although the child network obtained by NAS-RL algorithm reaches the SOTA, the number of hyperparameters that constitute the layer structure is so large that the search space becomes enormous. In addition, during the initialization, the RNN is generated randomly for the child network without adding any prior knowledge, which may lead to the subsequent child network obtained by adjusting based on this network does not have a better performance.

\subsubsection{Transformer in Neural Architecture Search}

 As shown above, the GPT model is taken from the decoder of the Transformer, and there are several current efforts to apply the Transformer to NAS. 

Ding \textit{et al.} \cite{ding2021hr} designed lightweight Transformers whose complexity can be dynamically adjusted according to different objective functions as well as computational budgets, and then encoded it together with convolution into a high-resolution search space to model. Since Transformers is designed for NLP tasks and is not optimal if applied directly to image processing tasks, Chen \textit{et al.} \cite{chen2021glit} designed the locality module to achieve a balance between global and local information. In addition, for the problem of huge search space, a hierarchical neural architecture search method is also proposed to search for the optimal visual transformer from two levels separately using an evolutionary algorithm.

The major difference between the above studies and the research in this paper is that the above studies design the Transformer as a module and put it into the search space. However, the work in this paper does not introduce the GPT model into the search space but uses the GPT model to enhance the search performance without affecting the original search strategy. In addition, the GPT model is obtained by training the neural architecture as input.


\section{Proposed Method}
\label{Proposed Method}


\subsection{Problem Setting}
NAS is a technique for automating the design of neural architectures. It uses various search strategies to explore a predefined search space and find the best-performing architecture for a given task. However, the huge search space makes the search for the optimal solution a rugged process. To this end, we concentrate on optimizing the search process and the search space.

The neural architecture consists of operations designed in advance. For operation $\left\{op_m\right\}$ ($m$ denotes the number of operations), it generally includes layers (\textit{e.g.}, convolution, pooling), blocks or cells (made up of multiple layers), etc. By permuting operations, we can obtain the search space $\mathcal{X} = \left\{x_1, x_2, x_3, \dots, x_n\right\}$ ($n$ denotes the number of architectures). However, the value of $n$ is normally huge. For example, if one wants to design a neural architecture with a depth of 20, then the total amount of possible neural architectures in the search space is $m^{20}$. The purpose of NAS is to find the optimal network architecture from the search space, \textit{i.e.}, $x^* = f(\mathcal{X})$, where $x^*$ denotes the optimal architecture and $f(\cdot)$ denotes the search strategy.

The motive of this paper is to optimize the search process and search space so as to enhance the ability to search for the optimal solution. In pursuit of this, we aim to introduce prior knowledge into the search process of neural architectures. Therefore, the problem in this paper is how to effectively introduce prior knowledge into the search process, which can be formulated as:

\begin{equation}
\label{problem_eq}
    x^*=f(\mathcal{X}, \mathbf{G}),
\end{equation}
where $\mathbf{G}$ denotes prior knowledge introduced.

\subsection{Algorithm Overview}

To effectively introduce $\mathbf{G}$ (as shown in Eq. \ref{problem_eq}), we propose the framework GPT-NAS, which uses the GPT model to optimize the NAS algorithm. It is widely accepted that the pre-trained GPT model is extremely gifted at predicting text, and this study takes advantage of it. By training the GPT model on large-scale neural architectures, the goal is to build a general understanding of neural architectures and transfer it to specific tasks. Specifically, we divide the content of the framework into three procedures, \textit{i.e.}, neural architecture encoding, pre-training and fine-tuning the GPT model, and neural architecture search. The details are described below:

\textbf{Neural Architecture Encoding.} In order for the GPT model to recognize neural architectures, it is necessary to encode them. The encoding strategy translates the neural architecture into the form of characters, where each character corresponds to a specific operation in the architecture, such as a convolutional layer, a fully connected layer, etc.

\textbf{Pre-Training and Fine-Tuning the GPT Model.} Pre-training and fine-tuning are two critical procedures for developing and deploying high-performance GPT models. Let the GPT model be pre-trained on a large-scale neural architecture dataset to achieve a general understanding of the neural network and fine-tuned it in specific tasks.

\textbf{Neural Architecture Search.} The neural architecture search procedure consists of two parts, namely architecture search and reconstruction. For the former, the architecture is mainly sampled, trained and evaluated using GA as the search strategy. For the latter, the sampled architectures are reorganized using GPT.

The core of the GPT-NAS framework, consisting of the three producers described above, lies in the optimization of the architecture obtained from the search using the GPT model. Through continuous iteration, the optimal architecture is found.

\subsection{Neural Architecture Encoding}
\label{Neural Architecture Encoding}

\begin{table*}[t]
\centering
\caption{The properties of different layers for describing}
\label{property}
\centering
\resizebox{\textwidth}{!}{
\centering
\begin{tabular}{ccccccc} 
\hline
\textbf{No.} & \textbf{Name} & \textbf{Conv} & \textbf{Pooling} & \textbf{FC} & \textbf{Other} & \textbf{Remark}                               \\ 
\hline
1            & id            & \checkmark             & \checkmark                & \checkmark           & \checkmark              & an identifier with an integer value           \\
2            & type          &               & \checkmark                &             &                & a string value                                \\
3            & name          &               &                  &             & \checkmark              & a string value                                \\
4            & in\_size      & \checkmark             & \checkmark                & \checkmark           & \checkmark              & input size of a three-element integer tuple   \\
5            & out\_size     & \checkmark             & \checkmark                & \checkmark           & \checkmark              & output size of a three-element integer tuple  \\
6            & kernel        & \checkmark             & \checkmark                &             &                & a two-element integer tuple                   \\
7            & stride        & \checkmark             & \checkmark                &             &                & a two-element integer tuple                   \\
8            & padding       & \checkmark             & \checkmark                &             &                & a four-element integer tuple                  \\
9            & dilation      & \checkmark             & \checkmark                &             &                & an integer                                    \\
10           & groups        & \checkmark             &                  &             &                & an integer                                    \\
11           & value         &               &                  &             & \checkmark              & a tuple                                       \\
12           & bias\_used    & \checkmark             & \checkmark                &             &                & a boolean number                              \\
\hline
\end{tabular}}
\end{table*}

Effective encoding of neural architectures can facilitate the GPT model to learn the fundamental laws of architectural composition. As a result, designing a general encoding strategy to accommodate the popular CNN architecture is necessary.


Based on the characteristics of the CNN architecture, we divide the structures that make up the CNN architecture into four categories \cite{sun2021arctext}: convolutional layer, pooling layer, fully connected layer, and other layers. Among them, the first three structures are necessary for almost CNN architectures, and the last one is used to represent all the remaining structures not included in the first three, such as the activation function, Batch Normalization (BN) \cite{luo2019strong}, etc.

\begin{enumerate}[(1)]
\item \textbf{Convolutional Layer}: The convolutional layer is the most fundamental and vital structure in CNNs. In convolutional operations, the core technology is the use of convolutional kernels (filter), which aim to extract features from the input image. The convolution kernel is a two-dimensional matrix (corresponding to height and width) and the parameters can be learned. In addition, the convolution kernel slides in the horizontal and vertical directions of the original image according to the \textit{stride}, respectively.  In general, regardless of the value of \textit{stride} (when the size of convolution kernel is not 1), the newly obtained feature map will certainly be smaller than the size of the original image, and this is also not conducive to the edge information of the image to work (because the convolution kernel will compute the central region of the image several times, while the edge region is relatively small). Therefore, the surrounding \textit{padding} of the image is needed to solve the above problem. Commonly, the convolution kernel convolves on each channel of the input feature map, which is a channel dense connection. In contrast, there is a channel sparse connection, which groups the input feature map channels and convolves each group separately. This process is called \textit{groups} and has the advantage of effectively reducing the number of parameters. In most cases, the convolutional kernel size needs to be enlarged if the respective field of a larger feature map is desired. However, the consequent drawback is that the number of parameters increases, so \textit{dilation} appears, which is an operation that injects space into the standard convolution kernel. In conclusion, the properties of the convolution layer are input size, output size, convolution kernel size, stride size, the number and value of padding, the space size of the kernel, the number of groups for the channels in the input feature map, and whether to use bias term.

\item \textbf{Pooling Laye}r: The property of the pooling layer are very similar to those of the convolutional layer, except that the following details need to be changed. First, the pooling layer contains two types, \textit{i.e.}, max pooling (\textit{MAX}) and average pooling (\textit{AVG}), so the property \textit{type} needs to be introduced to indicate which type is chosen. Second, the purpose of pooling is to reduce the size of the feature map, but this process has no parameters to learn, so there is no need for property \textit{groups} to reduce the number of parameters. Third, as we all know, the purpose of \textit{padding} is to expand the values in all four directions of the feature map, so not only the values but also the quantities need to be defined. However, in the pooling layer, the value of \textit{padding} is not set manually but the default value of 0. If not, it will affect the selection of feature values and make the final result biased. In summary, compared with the convolutional layer, the property \textit{groups} is removed and a new property \textit{type} is added. In addition, the value of \textit{padding} is also changed, so the pooling layer still maintains ten properties.

\item \textbf{Fully-Connected Layer}: Compared to the above two structures of the network, the fully-connected layer is straightforward to express. The fully-connected layer has only two properties, \textit{i.e.}, \textit{in\_size} and \textit{out\_size}. In many neural architectures for vision-related tasks, a fully-connected layer is necessary, such as in image classification tasks, where the final result is output for classification only through a fully-connected layer. Therefore, the structure is simple but essential.

\item \textbf{Other Layers}: In the CNN architecture, there are also many structures with different functions, such as activation function, BN, etc. These structures play the role of catalysts in the CNN architecture and enhance the performance of the neural architecture. Therefore, in this part, all relevant structures will be described. The common properties of these structures are the \textit{name}, \textit{in\_size}, \textit{out\_size}, and \textit{value}. Among them, \textit{value} denotes the relevant parameter involved in the corresponding structure
\end{enumerate}

As shown above, we have described the properties of different structures in CNN, and these properties will represent the textual data form of a CNN architecture. The specific information is shown in Table \ref{property}.


\subsection{Pre-Training and Fine-Tuning the GPT Model}
\label{Pre-Training and Fine-Tuning the GPT Model}

Providing the GPT model with a general understanding of the fundamental laws of neural architecture is the heart of this study. Therefore pre-training and fine-tuning the GPT model using the neural architecture as training data can effectively achieve what is needed in this paper, \textit{i.e.}, introducing prior knowledge in the neural architecture search process.

The pre-training phase of a GPT model requires training on a large amount of data to learn the parameter distribution, followed by a fine-tuning phase to suit various tasks. As a result, the mainstream approach will use unsupervised learning for maximum likelihood estimation in the first phase and use supervised learning to optimize the model using a cross-entropy loss function in the second phase. However, in this study, since both the pre-training and fine-tuning phases are trained on the neural architectural dataset and ground truth is presented, it will be a better choice to use a supervised learning approach for both phases.

In the proposed GPT-NAS, we leverage the GPT model to predict the next data by the previous information, \textit{i.e.} the structural information of the previous layers to predict the structural information of the next layer (see Section \ref{Neural Architecture Search} for details). Thus, based on the given layer structures, we will minimize the following objective function:

\begin{equation}
    \begin{cases}
    \begin{matrix}
    {\mathcal{F} = {\sum_{t}^{T}{\mathcal{L}\left( \hat{l}_{t}, l_t \middle| l_{< t},C \right)}}} \\
    \end{matrix} \\
    \begin{matrix}
    {C = f\left( \theta,\mathcal{S} \right)} \\
    \end{matrix} 
\end{cases}
\end{equation}
where \textit{C} denotes the neural architecture, $\theta$ is the parameters used in constituting the neural architecture, and $\mathcal{S}$ is the corresponding network layer structures. In the objective function $\mathcal{F}$, \textit{T} denotes the amounts of layers in \textit{C}, $\mathcal{L}$ denotes the loss function, $\hat{l}_{t}$ and $l_t$ denote the predicted layer structure and the true layer structure obtained at layer $t$, respectively, and $l_{<t}=(l_{t-k},…,l_{t-1})$ (\textit{k} is the size of the data window).

\subsection{Neural Architecture Search}
\label{Neural Architecture Search}

\begin{figure*}[t]
	\begin{center}
		\includegraphics[width=0.95\linewidth]{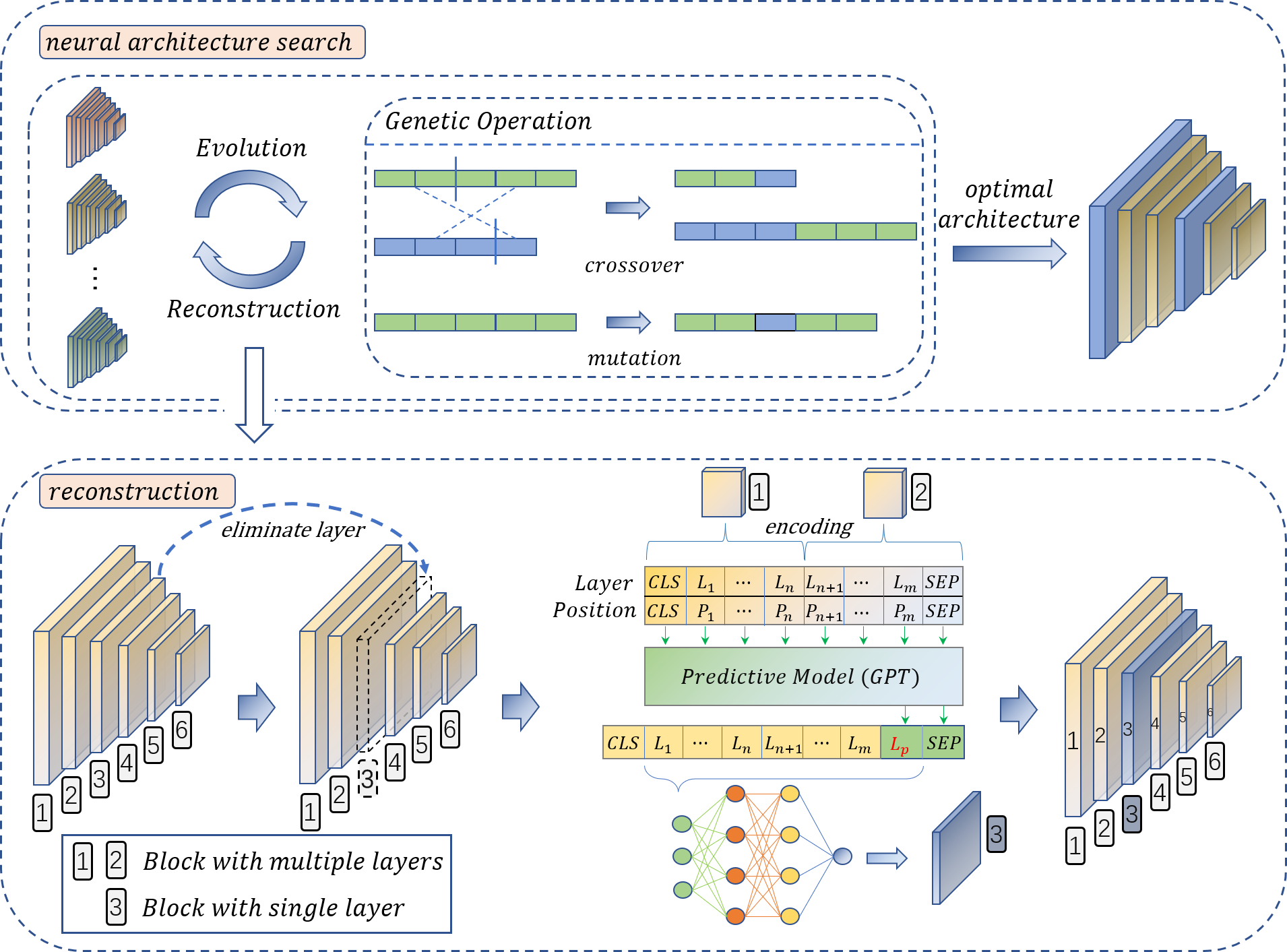}
	\end{center}
	\caption{The flowchart of neural architecture search. We divide this procedure into two parts, \textit{i.e.}, search and reconstruction. The former is a generalised neural architecture search method using GA as the search strategy. The latter is a reorganisation of the structure of the neural architecture by using the GPT model. For example, for a neural architecture, we eliminate the third block with a certain probability and then use the GPT model to re-predict the third block structure based on the information from the first two blocks.}
	\label{GPT_NAS}
\end{figure*}

After the pre-trained and fine-tuned GPT model is obtained, we next describe how it can be introduced into the process of neural architecture search.

The procedure of neural architecture search consists of two main parts, \textit{i.e.}, network architecture search and architecture reconstruction, and Fig. \ref{GPT_NAS} shows the flowchart. 

\begin{algorithm}
	\caption{Structure Prediction}\label{Structure Prediction}
	\KwIn{The neural architecture with eliminated structures \textit{cnn\_ori}}
	\KwOut{The neural architecture with optimization \textit{cnn\_new}}
	$\textit{cnn\_new} \leftarrow \textit{null}$;\\
	\For{\textit{index}, \textit{layer} in \textit{cnn\_ori}}
	{\uIf{\textit{layer} is eliminated}{
	$\textit{data} \leftarrow$ transform $layer_i$ into textual data ($i \in [0, ..., index-1]$);\\
	$new\_layer \leftarrow$ prediction layer using the GPT based on data;\\
	$new\_block \leftarrow$ prediction block using the FCN based on \textit{data} and \textit{new\_layer};\\
	$cnn\_new \leftarrow cnn\_new \cup new\_block$;\\
	}
	\Else{
	$cnn\_new \leftarrow cnn\_new \cup \textit{layer}$;\\
	}
	}
	\textbf{Return} $cnn\_new$.
\end{algorithm}

\begin{enumerate}[(1)]
\item \textbf{Network Architecture Search}: In the first part, we implement the search of neural architectures with an evolutionary algorithm as the search strategy. Specifically, multiple individuals are initialized randomly, and each individual is represented as a neural architecture. Note that a variable-length encoding strategy is used in this study, \textit{i.e.}, each neural architecture has a different depth. Then, each neural architecture is reconstructed and the performance of the reconstructed architecture is evaluated. After that, during the iterative process, the GA is used to perform evolutionary operations on individuals, including crossover, mutation, and selection strategies to facilitate the acquisition of better-performance individuals and form a new population. Finally, the optimal individual will be obtained after reaching the maximum number of iterations.

\item \textbf{Architecture Reconstruction}: In the second part, we will regenerate the blocks of the architecture obtained from the search. Note that in order to reduce search space, we will build the neural architecture based on blocks. In the following, we use a neural architecture as an example to illustrate the process of reconstruction. 
Firstly, the layers in the architecture are selected with a certain probability, then the block containing that layer is eliminated and a new block structure is predicted to refill the position of the eliminated block based on the previous information. 
As shown in the bottom half of Fig. \ref{GPT_NAS}, the third block in the architecture with a depth of six is eliminated, and then the fine-tuned GPT model is used to predict the block structure based on the first and second blocks (suppose the first and second blocks contain multiple layers, the third block contains one layer). 
Secondly, since the composition of the neural architecture is based on blocks and the predictions obtained by the GPT model is a layer, we introduce a Fully Connected Network (FCN) to select the fittest blocks according to the layers (the blocks are designed in advance and is shown in Table \ref{Network architecture dataset}), which is based on the possibility that the same type of layer structure will exist in different blocks.
Continuing with the example in Fig. \ref{GPT_NAS}, new layers are obtained through the GPT model and then combined with the previous information, \textit{i.e.}, the layers in the first two blocks, and feed together into the FCN to obtain the corresponding block structure with the operation of classification (treating the different blocks as different categories). The details can be seen in the Algorithm \ref{Structure Prediction}. 

\end{enumerate}

In summary, the neural architecture obtained from each iteration is optimized using the GPT model, and performance improvement is achieved by changing its structure. In this case, we propose the concept of elimination rate to determine whether a layer of the neural architecture is selected and whether the corresponding block is eliminated. However, according to the law of evolutionary algorithms, the quality of the offspring population will be better than that of the parent population. Therefore, as the iteration proceeds, the elimination rate of the structures will be linearly decreasing. Eq. \ref{equation_linear_decrease} shows the elimination rate at the \textit{t}-th iteration. 

\begin{equation}
\label{equation_linear_decrease}
    {rate_{t} = rate_{ori} - \frac{iter_{t}}{iter_{max}} \times rate_{ori}} 
\end{equation}
where $rate_{ori}$ indicates the initialized elimination rate, $iter_{t}$ and $iter_{max}$ donote the \textit{t}-th iteration and the number of iterations.

\subsection{Acceleration Strategy}
\label{Acceleration strategy}

One of the criticisms of NAS development has been the time-consuming problem. In a recent study, \cite{mellor2021neural} proposed that the performance of neural architectures can be evaluated without training. However, the method suffers from assumptions and does not guarantee that the final results are the same as the real ones. Therefore, an adequate search for each architecture remains the dominant approach.

In this study, we introduce a new acceleration strategy to reduce the time cost of the GPT-NAS, as shown below.

\subsubsection{Only the structures obtained from the prediction are trained}

Training the entire neural architecture is time-consuming, but it is more efficient if only the ‘vital’ structures of the architecture are trained. In order to speed up the training while ensuring that the performance of the neural architecture is not lost, we assume that the layer structures obtained from the GPT model predictions are ‘vital’ structures and propose to train only these structures. (confirmed in Section \ref{Validation on acceleration strategies}). 
And for neural architectures without predicted structures, inspired by \cite{chen2021bn}, we will train all batch normalization (BN) in the neural architecture. Finally, if both of the above rules are not satisfied, the overall neural architecture is trained. 

\subsubsection{Only a small number of epochs are trained}
This strategy has been covered in some works \cite{zoph2018learning, baker2017accelerating}. While in this study, a smaller number of epochs will be used. The rationale for doing so is motivated by warmup \cite{he2016deep}, which makes the learning rate increase in fewer epochs to alleviate the model overfitting phenomenon and reach an equilibrium state. If the model stabilizes faster (\textit{i.e.}, the higher accuracy rate achieved) during this time, it can be considered a better performance of the model (confirmed in Section \ref{Validation on acceleration strategies}).

\section{Experimental Design}
\label{Experimental Design}
In order to verify the effectiveness of the proposed algorithm, we will conduct a series of experiments. Therefore, this section will present the design of all the elements involved in the experiments. First, we will introduce the state-of-the-art algorithm for comparison with GPT-NAS. Then, the datasets used in this experiment and the hyper-parameters are introduced.

\subsection{Peer Competitors}

To verify the effectiveness of the proposed algorithm, we selected several state-of-the-art algorithms for comparison in our experiments. We divide the selected peer competitors into two categories: algorithms obtained by manual design and those obtained by automatic search. Specifically, there are seven manually designed neural network architectures, namely EfficientNet-B0 \cite{tan2019efficientnet}, GoogLeNet \cite{szegedy2015going}, RegNet \cite{radosavovic2020designing}, ResNet-101 \cite{he2016deep}, ResNeXt-101 \cite{xie2017aggregated}, Shufflenet \cite{ma2018shufflenet}, and Wide-ResNet \cite{zagoruyko2016wide}. These neural architectures are chosen for two reasons. One is that they are very popular and representative in the vision domain, and the other is that the constituent blocks of the neural architecture of the search space for this experiment are extracted from these architectures, as described in Section \ref{dataset}. In the second category, we choose thirteen NAS algorithms based on different strategies to verify the superiority of GPT-NAS, and the corresponding algorithms are shown in Table \ref{all performance}.

\subsection{Datasets}
\label{dataset}

Since there are two parts of work in this study, \textit{i.e.}, the implementation of the GPT-NAS and the training of the GPT model, two types of datasets are required. Firstly, image datasets are needed for training the neural architecture, so the three most popular datasets are used here, namely CIFAR-10, CIFAR-100 \cite{krizhevsky2009learning}, and ImageNet-1K \cite{ILSVRC15}. Second, the neural architecture dataset is required in the training of the GPT model, especially in the pre-training phase, which requires a very large amount of data. Therefore, in the pre-training phase, we use NAS-Bench-101 \cite{ying2019bench}, while in the fine-tuning phase, the required dataset is randomly taken from the state-of-the-art neural architectures.

CIFAR-10 and CIFAR-100 are the two most popular image classification datasets. Each contains 60,000 images, of which 50,000 are used for training, and 10,000 are used for testing. The difference between the two is the number of object classes. On CIFAR-10, 5,000 images per category are used for training, while on CIFAR-100, only 500 images are used for training. Furthermore, ImageNet-1K is a more challenging dataset than the previous two, which has 1,000 object classes and contains 1,281,167 training images, 50,000 validation images, and 100,000 test images. Since the image data in the test set does not give the corresponding label, only the training and validation sets are used in this experiment. In addition, it should be noted that the ImageNet-1K is too large to be realistically applied to NAS, and there is no good method to deal with it currently. So, we only take 10\% of the training images to search the architecture (the validation set is consistent with the original data), but for the optimal architecture obtained from the search, we still use the whole dataset for training.

NAS-Bench-101 is a dataset of different neural architectures obtained by changing the structure of cell in a fixed framework. Each cell has at most seven vertices and nine edges, and each neural architecture is obtained by stacking randomly composed cell structures. In the dataset, there are 423,624 neural architectures, and the corresponding performance is obtained for multiple runs on the CIFAR-10. To make the GPT model learn better neural architectures in the pre-training phase, we do not select all the neural architectures, but those with the classification accuracy of 90\% or more on the validation set from them as the training data and the final amount of neural architectures is 295,889. The dataset in the fine-tuning phase adopts the most commonly used neural architectures nowadays. In Table \ref{Network architecture dataset}, we list the seven neural architectures and the corresponding blocks (the four blocks listed in the eighth row are those common to the neural architectures mentioned above). The number after each neural architecture in the table indicates the number of variants we can extend, depending on the properties of that neural architecture, for example, ResNet can have 18 layers, 34 layers, etc. So the total number of neural architectures is 36. Note that although the number of neural architectures in the pre-training phase is 295889 and the number of neural architectures in the fine-tuning phase is 36, when we train the GPT model, the input data is in units of layers, not architectures, and thus the resulting training samples far exceed the number of neural architectures. In addition, in Table \ref{Network architecture dataset}, we extract 15 blocks based on the different neural architectures that are used to compose the neural architectures in the search space.

\begin{table}
\centering
\caption{Neural architecture dataset for fine-tuning phase}
\label{Network architecture dataset}
\centering
\setlength{\tabcolsep}{6mm}{
\begin{tabular}{ccc} 
\hline
\textbf{id}        & \textbf{Neural Architecture}    & \textbf{Block}      \\ 
\hline
\multirow{2}{*}{1} & \multirow{2}{*}{EfficientNet(8)} & ConvNormActivation  \\
                   &                                  & SqueezeExcitation   \\
\hline
\multirow{2}{*}{2} & \multirow{2}{*}{GoogleNet(1)}    & Inception           \\
                   &                                  & Avgpool             \\
\hline
\multirow{2}{*}{3} & \multirow{2}{*}{RegNet(14)}      & ResBottleneckBlock  \\
                   &                                  & Stem                \\
\hline
\multirow{2}{*}{4} & \multirow{2}{*}{ResNet(5)}       & Bottleneck          \\
                   &                                  & Basicblock          \\
\hline
5                  & ResNext(2)                       & Bottleneck          \\
\hline
6                  & ShuffleNet(4)                    & InvertedResidual    \\
\hline
7                  & wide-ResNet(2)                   & Bottleneck          \\
\hline
\multirow{4}{*}{8} & \multirow{4}{*}{other}           & Maxpool             \\
                   &                                  & BatchNormal         \\
                   &                                  & Relu                \\
                   &                                  & Conv                \\
\hline
\end{tabular}}
\end{table}

\subsection{Parameters Settings}

The parameter settings in the experiment can be divided into two parts, one for GPT-NAS and the other for the GPT model. In the following, we will describe in detail the parameters involved in these two parts. 

In GPT-NAS, the most critical element is eliminating network structures from each neural architecture and applying the GPT model to predict and refill the eliminated network structures. Therefore, we determine the initial elimination rate of network structures to be 0.4 in advance in this experiment (confirmed in Section \ref{Validation on different elimination rates}). Second, for the depth of neural architectures, we set the number of blocks in the range [10, 20]. Then, since GPT-NAS is optimized based on GA, we set the size of populations to 30, the number of iterations to 20, and the crossover and mutation rates to 0.9 and 0.3, respectively. Finally, for the neural architecture training, we made the following settings for the parameters in it. Specifically, we set the number of epochs to 6, and use stochastic gradient descent (SGD) \cite{lecun1998gradient} to optimize the parameters, while the learning rate is linearly incremented to 0.01. In addition, due to the different image datasets, we set the batch size differently. On CIFAR-10 or CIFAR-100, the batch size is 512, while ImageNet-1K is 128. When the running is finished, the neural architecture with optimal accuracy will be obtained and retrained. On CIFAR-10 and CIFAR-100, the optimal neural architecture is trained for 350 epochs, and the learning rate decays to 1/10 of the original rate every 100 epochs starting from 0.01. While on ImageNet-1K, we only train the optimal architecture for 120 epochs due to resource constraints and the learning rate decays every 30 epochs.

For the GPT model, we mostly used the same parameter settings as in the seminal paper, with a few differences as shown below. As we all know, the GPT model largely followed \cite{vaswani2017attention} and trained a 12-layer decoder-only transformer. However, in our experiments, since the amount of data is not as large as in the task of the seminal paper, only 4 layers of decoders are trained and only 4 attention heads are introduced in each decoder. After collation, 168 different network layer structures are finally obtained. In addition, we set the input dimension to 10 and the stride to 1. Finally, we trained the model for 300 epochs using the Adam optimizer with a learning rate of 1e-4 and a batch size of 128.

\section{Experimental Results}
\label{Experimental Results}

In this section, we will discuss the experimental results in detail. The analysis of the experiments is divided into two parts, the first part is the overall performance comparison of the proposed algorithm with other state-of-the-art algorithms (Section \ref{Performance Overview}), and the second part is the ablation experiment (Section \ref{Ablation Experiments})

\subsection{Performance Overview}
\label{Performance Overview}

\begin{table*}[t]
\centering
\caption{Experimental results of the proposed algorithm and the state-of-the-art algorithm on different datasets.}
\label{all performance}
\centering
\resizebox{\textwidth}{!}{
\centering
\begin{tabular}{cccccccccc} 
\hline
\multirow{2}{*}{Search Method} & \multirow{2}{*}{Architectures} & CIFAR-10   & CIFAR-100  & \multirow{2}{*}{Param(M)} & \multirow{2}{*}{GPU Days} & \multicolumn{2}{l}{ImageNet-1K} & \multirow{2}{*}{Param(M)} & \multirow{2}{*}{GPU Days}  \\ 
\cline{3-4}\cline{7-8}
                               &                                & Top1 (\%) & Top1 (\%) &                           &                           & Top1 (\%) & Top5 (\%)           &                           &                            \\ 
\hline
\multirow{7}{*}{Human}         & EfficientNet-B0 \cite{tan2019efficientnet}                & 97        & \textbf{86.6}      & 5.3                       &                           & 77.69     & 93.53               & 5.3                       &                            \\
                               & GoogLeNet \cite{szegedy2015going}                      & 89.23     & 62.9      & 6.6                       &                           & 69.78     & 89.53               & 6.6                       &                            \\
                               & RegNet \cite{radosavovic2020designing}                         & 92.72     & 70.19     & 31.3                      &                           & 76.57     & 93.07               & 31.3                      &                            \\
                               & ResNet-101 \cite{he2016deep}                     & 93.57     & 74.84     & 44.5                      &                           & 77.37     & 93.55               & 44.5                      &                            \\
                               & ResNeXt-101 \cite{xie2017aggregated}                    & 96.29     & 82.27     & 18.1                      &                           & 77.8      & 94.3                & 18.1                      &                            \\
                               & ShuffleNet \cite{ma2018shufflenet}                     & 90.87     & 77.14     & 3.5                       &                           & 73.7      & 91.09               & 3.5                       &                            \\
                               & Wide-ResNet \cite{zagoruyko2016wide}                    & 95.83     & 79.5      & 36.5                      &                           & 78.1      & 93.97               & 68.9                      &                            \\ 
\hline
\multirow{5}{*}{RL}            & NAS-RL \cite{zoph2016neural}                         & 96.35     & N/A       & 37.4                      & 22,400                     & N/A       & N/A                 & N/A                       & N/A                        \\
                               & MetaQNN \cite{baker2016designing}                        & 93.08     & 72.86     & 11.2                      & 100                       & N/A       & N/A                 & N/A                       & N/A                        \\
                               & EAS \cite{cai2018efficient}                            & 95.77     & N/A       & 23.4                      & 10                        & N/A       & N/A                 & N/A                       & N/A                        \\
                               & NASNet-A \cite{zoph2018learning}                       & 96.59     & N/A       & 3.3                       & 2,000                      & 74        & 91.6                & 5.3                       & 2,000                       \\
                               & Block-QNN-S \cite{zhong2018practical}                    & 96.46     & 81.94     & 39.8                      & 96                        & 77.4      & 93.5                & N/A                       & 96                         \\ 
\hline
\multirow{4}{*}{EA}            & Large-scale Evo \cite{real2017large}                & 94.6      & 77        & 5.4/40.4                  & 2,750                      & N/A       & N/A                 & N/A                       & N/A                        \\
                               & GeCNN \cite{xie2017genetic}                          & 94.61     & 74.88     & N/A                       & 17                        & 72.13     & 90.26               & 156                       & 17                         \\
                               & AE-CNN \cite{sun2019completely}                         & 95.3      & 77.6      & 2/5.4                     & 27/36                     & N/A       & N/A                 & N/A                       & N/A                        \\
                               & GPCNN \cite{suganuma2017genetic}                          & 94.02     & N/A       & 1.7                       & 27                        & N/A       & N/A                 & N/A                       & N/A                        \\ 
\hline
\multirow{4}{*}{GO}            & SNAS \cite{xie2018snas}                           & 97.17     & 82.45     & 2.8                       & 1.5                       & 72.7      & 90.8                & 4.3                       & 1.5                        \\
                               & P-DARTS \cite{chen2019progressive}                        & 97.33     & N/A       & 3.51                      & 0.3                       & 75.3      & 92.5                & 5.1                       & 0.3                        \\
                               & DARTS \cite{liu2018darts}                          & 97.14     & 82.46     & 3.4                       & 0.4                       & 76.2      & 93                  & 4.9                       & 4.5                        \\
                               & ISTA-NAS \cite{yang2020ista}                       & 97.64     & N/A       & 3.37                      & 2.3                       & 76        & 92.9                & 5.65                      & 33.6                       \\ 
\hline
\multirow{2}{*}{Ours}          & NAS without GPT                       & 90.77     & 75.20           & 4.6/38.05                       & 1.5                       & 69.53     &                              & 104.67                    & 4                          \\
                               & NAS with GPT (GPT-NAS)                       & \textbf{97.69}     & 82.81           & 7.1/10.5                       & 1.5                       & \textbf{79.08}     & \textbf{95.92}               & 110.94                    & 4                          \\
\hline
                               &                                &           &           &                           &                           &           &                     &                           &                           
\end{tabular}}
\end{table*}

In this section, we will describe the results of comparing GPT-NAS with other algorithms, and the specific experimental results are shown in Table \ref{all performance}. In the experiments, GPT-NAS is compared with four categories of related neural architectures. In addition, we also list the accuracy of the optimal neural architecture with and without the GPT model. The table shows the experimental results of different algorithms on different datasets. Two points should be noted here. The first is that there are no GPU Days (a metric used to measure the time cost of NAS-related algorithms) for the neural architecture obtained by manual design, and the second is that `N/A' denotes null values. The reason for the null value is that most of the experimental results are taken from the seminal paper of the corresponding algorithm. If the original authors do not test this dataset, we will use 'N/A' to indicate it. In addition, on CIFAR-10 and CIFAR-100, only Top1 is selected as the final accuracy due to the small number of categories, while on ImageNet-1K, both Top1 and Top5 metrics are selected as the final accuracy representation. The best results on each dataset have been marked in bold.

On CIFAR-10, the neural architecture obtained by GPT-NAS achieves the best result among all algorithms, with 97.69\%. Compared to the manually designed neural architecture, GPT-NAS improves the classification accuracy by up to nearly 9\%. Furthermore, the accuracy has increased by 4.12\% compared to the ResNet-101, which is the most famous architecture today. Moreover, among the remaining architectures, GPT-NAS is 0.69\% higher than EfficientNet-B0, which is the smallest accuracy difference, and 4.97\%, 1.4\%, 6.82\%, and 1.86\% higher than RegNet, ResNeXt-101, ShuffleNet, and Wide-ResNet, respectively. The better performance of GPT-NAS over the manually obtained neural architecture reflects that the neural architecture composed of different blocks is efficient and also demonstrates that the neural architecture learns global information. After comparing with NAS algorithms based on different strategies, it can be found that GPT-NAS also has the best performance, which is 4.61\% higher than MetaQNN. Moreover, GPT-NAS is an algorithm based on EA, when compared with four listed state-of-the-art EA-NAS algorithms, it still outperforms more than 2\% of them. Among all the algorithms involved in the comparison, the GO-based algorithm has the best average performance, all above 97\%, but GPT-NAS still has a slight edge. 

On CIFAR-100, GPT-NAS is second only to EfficientNet-B0 among all algorithms. Compared to CIFAR-10, CIFAR-100 is significantly more challenging, with only five of all the algorithms involved in the comparison exceeding 80\% in accuracy, while none of the EA-NAS algorithms exceeds 78\%. Although the accuracy of GPT-NAS is lower than EfficientNet-B0, its advantage is still undeniable compared with other algorithms. For example, it is about 20\% higher than GoogLeNet. Furthermore, compared with other EA-NAS algorithms in the same category, GPT-NAS is the only one with an accuracy of more than 80\%. On the other hand, regarding the number of parameters, the neural architecture obtained by searching on either CIFAR-10 or CIFAR-100 is not significantly superior. However, it is also in the middle to the upper level. As for GPU Days, GPT-NAS can be ranked in the top three, and this comparison is a qualitative improvement over the RL-based and EA-based algorithms. Note that most of the GO-based algorithms use the method of constructing a supernet, and the subsequent subnetworks implement the weight sharing, so the time consumption is significantly reduced, which is different from training all neural architectures in this paper.

Finally, GPT-NAS outperforms all other algorithms on ImageNet-1K and is the only algorithm with a classification accuracy of over 79\% on Top1 and over 95\% on Top5. Since ImageNet-1K is more difficult to classify, fewer algorithms are involved in the comparison, while the classification accuracy does not differ significantly among algorithms. Except for GoogLeNet, all other algorithms have accuracies between 72\% and 78\% on Top1. Among the manually designed neural architectures, the optimal one is Wide-ResNet with an accuracy of 78.1\%, which is 0.98\% lower than GPT-NAS, while among the NAS algorithms, the optimal one is Block-QNN-S, but its accuracy is also 1.68\% lower than GPT-NAS. The only drawback is that the number of neural architecture parameters obtained by GPT-NAS is relatively large, only less than that of GeCNN. Finally, GPT-NAS can be ranked third on GPU Days, with a quarter reduction compared to GeCNN, an algorithm also based on EA.

On the other hand, by comparing the performance of the NAS in this experiment with and without the GPT model, we can find that the accuracy of the neural architecture obtained with the introduction of the GPT model is generally improved on all datasets. On the three datasets, the accuracy is improved by 7\%, 9\% and 12\% respectively, which demonstrates the effectiveness of our method.

\subsection{Ablation Experiments}
\label{Ablation Experiments}

For the method proposed in this experiment, two ablation experiments are performed to verify its effectiveness. In the first part, the influence of different elimination rates on the neural architecture is compared (Section \ref{Validation on different elimination rates}). In the second part, we will test the effectiveness of the proposed acceleration strategy (Section \ref{Validation on acceleration strategies}). 
Note that the parameter settings for the experiments in this section will be slightly different from those in the main experiment (Section \ref{Performance Overview}), as described in each subsection. In addition, ablation experiments are tested on both CIFAR-10 and CIFAR-100 datasets.

\subsubsection{Validation on different elimination rates}
\label{Validation on different elimination rates}

The core of this study is to eliminate the structures in the neural architecture effectively and to perform prediction and reconfiguration, so it is important to choose the appropriate elimination rate. In this subsection, we experiment with different elimination rates and choose the optimal one. For convenience, we only tested the initialized individuals and did not perform genetic operations. Specifically, we chose 15 initialized neural architectures and trained 90 epochs with elimination rates of 0, 0.2, 0.4, 0.6, and 0.8 to test their classification accuracy. When the elimination rate is 0, it means that the neural architecture has not changed its internal structure.
The experimental results are shown in Table \ref{table_different elimination rates}. 
In Table \ref{table_different elimination rates}, the ``mean value'' indicates the average accuracy of the 15 neural architectures at the corresponding elimination rates, while "+/=/-" indicates the number of individual neural architectures with elimination rates that are better, equal, and worse in terms of classification accuracy than those without elimination rates.

From Table \ref{table_different elimination rates}, we can obtain that on CIFAR-10, the effect is the worst when the elimination rate is 0.2, with five neural architectures worse than the case without elimination rate, and the next is when the elimination rate is 0.8, with two neural architectures worse than the case without elimination rate. In addition, on CIFAR-100, only when the elimination rate is 0.4, all neural architectures are better than the case without the elimination rate. Furthermore, from the metric of ``mean value'', we can find that the average accuracy of the neural architecture on CIFAR-10 is improved by at least 5\%, up to 14\%, compared to the case without elimination rate. While on CIFAR-100, the average accuracy of the initialized neural architecture is improved by at least 8\% after the introduction of the elimination rate. In summary, the neural architectures with the introduction of the elimination rate have a huge average performance improvement, especially with an elimination rate is 0.4. Therefore, in the main experiment, we chose the elimination rate of 0.4 for the network structure as the final criterion.


\begin{table}
\centering
\caption{Experimental results of neural architecture with different elimination rates.}
\label{table_different elimination rates}
\centering
\setlength{\tabcolsep}{4mm}{
\begin{tabular}{cccc} 
\hline
\textbf{Dataset}                  & \begin{tabular}[c]{@{}l@{}}\textbf{elimination rate}\end{tabular} & \textbf{mean value} & \textbf{+/=/-}   \\ 
\hline
\multirow{5}{*}{CIFAR-10}  & 0                                                                   & 0.3792     &         \\
                          & 0.2                                                                 & 0.4226     & 10/0/5  \\
                          & 0.4                                                                 & 0.5134     & 15/0/0  \\
                          & 0.6                                                                 & 0.5056     & 15/0/0  \\
                          & 0.8                                                                 & 0.4828     & 13/0/2  \\ 
\hline
\multirow{5}{*}{CIFAR-100} & 0                                                                   & 0.0453     &         \\
                          & 0.2                                                                 & 0.1257     & 14/1/0  \\
                          & 0.4                                                                 & 0.1392     & 15/0/0  \\
                          & 0.6                                                                 & 0.1278     & 13/0/2  \\
                          & 0.8                                                                 & 0.1261     & 1/4/10  \\
\hline
\end{tabular}}
\end{table}

\subsubsection{Validation on acceleration strategies}
\label{Validation on acceleration strategies}

\begin{figure}[t]
	\begin{center}
		\includegraphics[width=0.95\linewidth]{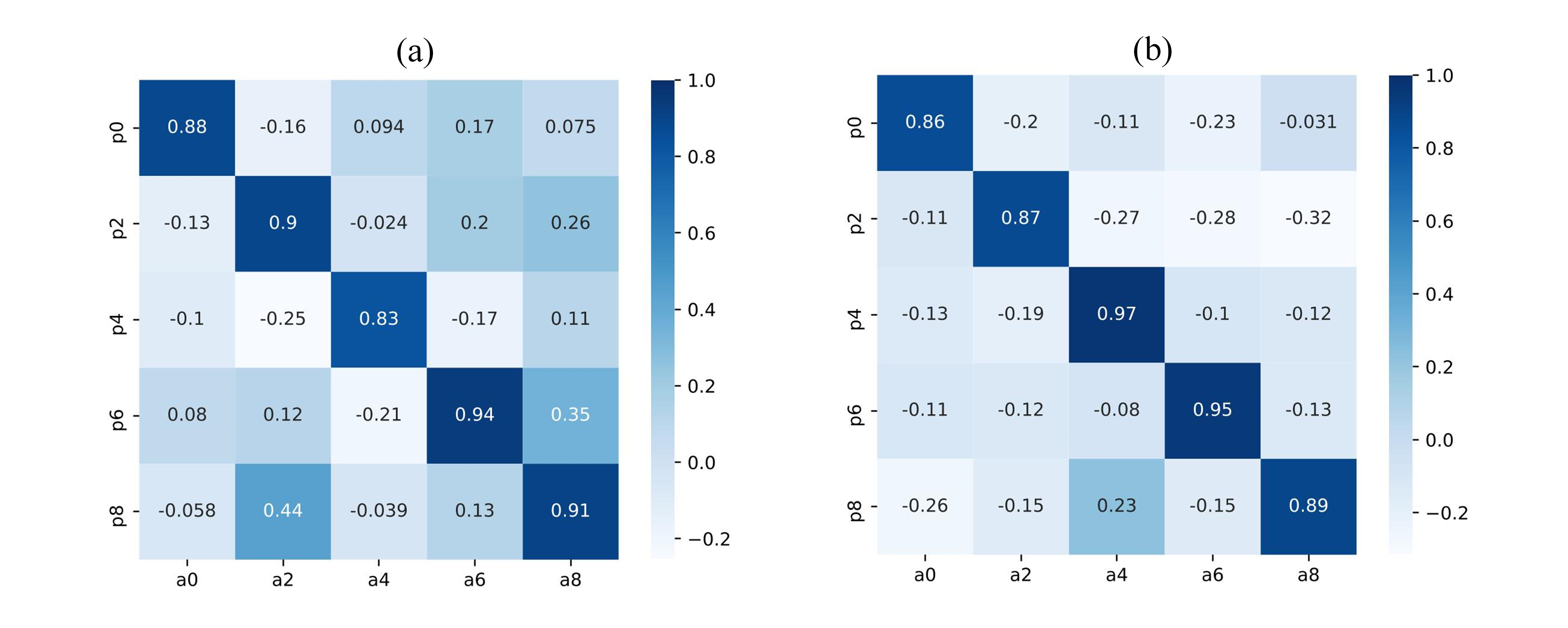}
	\end{center}
	\caption{Comparison of the accuracy correlation results achieved by training the blocks obtained by prediction and all blocks in the neural architecture respectively. (a): CIFAR-10; (b): CIFAR-100.}
	\label{accuracy correlation}
\end{figure}

\begin{table}
\centering
\caption{Comparison of the correlation between the accuracy achieved by the neural architecture trained on a small number of epochs versus multiple epochs.}
\label{table_epoch comparation}
\centering
\setlength{\tabcolsep}{5mm}{
\begin{tabular}{cccc} 
\hline
                        & \textbf{Dataset} & \textbf{\textit{Pearson}} & \textbf{\textit{p-value}}   \\ 
\hline
\multirow{2}{*}{$e_6-e_{30}$} & CIFAR-10  & 0.7247  & 2.68E-09  \\
                        & CIFAR-100 & 0.8657  & 4.96E-16  \\ 
\hline
\multirow{2}{*}{$e_6-e_{60}$} & CIFAR-10  & 0.6984  & 1.72E-08  \\
                        & CIFAR-100 & 0.8176  & 4.33E-13  \\ 
\hline
\multirow{2}{*}{$e_6-e_{90}$} & CIFAR-10  & 0.7046  & 1.13E-08  \\
                        & CIFAR-100 & 0.8107  & 9.66E-13  \\
\hline
\end{tabular}}
\end{table}


In this subsection, we will implement two main types of experiments. Firstly, the neural architecture optimized by the GPT model is trained in two parts, \textit{i.e.}, only on the predicted blocks and on all blocks in the neural architecture, and then the two correlations are calculated. Secondly, the neural architecture is trained under different numbers of epochs and the correlation between them is calculated. Note that for correlation comparison of accuracy, we mainly use the \textit{Pearson correlation coefficient} (PCC, between -1 and 1, the larger the value, the higher the correlation) and \textit{p-value} (less than 0.05 means that they are correlated) to show.

For the first experiment, Fig. \ref{accuracy correlation} shows the accuracy correlation heat map of the neural architecture for training only the blocks obtained by prediction and all blocks. The horizontal axis indicates that all blocks are trained, while the vertical axis indicates that the predicted blocks are trained. For example, ``a2'' means that the neural architecture is optimized with the elimination rate of 0.2 and all blocks are trained, while "p2" means that only the predicted blocks are trained. In addition, the experimental results in the figure are obtained by averaging the PCC calculated by each of the 15 neural architectures.  For the second experiment, Table \ref{table_epoch comparation} gives the experimental results on whether there is a correlation between training only a small number of epochs versus training multiple epochs. In addition to the PCC, we also list the corresponding \textit{p-values}. In addition, the first column of the table indicates the comparison between different epochs, for example, ``$e_6-e_{30}$'' indicates whether the accuracy values obtained from the two pieces of training are linearly correlated in the case of 6 epochs and the case of 30 epochs. 

As can be obtained from Fig. \ref{accuracy correlation}, the color on the diagonal in the heat map is the darkest regardless of the dataset, which means that the corresponding correlation is the highest and the values are above 80\%. And the values on each diagonal line indicate the accuracy of training only the predicted blocks is relevant to training all blocks on the same neural architecture. In addition, in Table \ref{table_epoch comparation}, we can find that the effect of using a small number of epochs to train is the same as that of using most epochs to train. It is expressed as a positive correlation on PCC, while a linear correlation between them can be proved by \textit{p-value}.

\section{Conclusion and Future work}
\label{Conclusion and Future work}

NAS algorithm automates the design of neural networks, but its large search space makes finding the optimal architecture challenging and time-consuming. In this context, we propose a novel approach called the GPT-NAS, which leverages the power of the GPT model and GA to guide the search process of the NAS algorithm to achieve the effect of reducing the search space. 
Specifically, we divide the algorithm into three procedures, namely, neural network coding, GPT model pre-training and fine-tuning, and neural architecture search. 
First, we encode the neural architectures into vector form for recognition by the GPT model. 
Second, the GPT model is pre-trained on the NAS-Bench-101 dataset and fine-tuned on a small number of neural architectures extracted from the search space for a specific task. 
The pre-training process provides the GPT model with a priori knowledge of the neural architecture, while the fine-tuning process allows for a tighter connection to the task at hand. 
Finally, the fine-tuned network architecture is introduced into the NAS algorithm with GA as the search streategy to guide the search. The proposed GPT-NAS is compared with 20 state-of-the-art competitors on three popular datasets, where the competitors contain manually designed algorithms and NAS algorithms. The analysis of the experimental results shows that the GPT-NAS achieves state-of-the-art results and proves that the GPT model has a boosting effect on the algorithm. In future work, we will study the GPT model in more depth to make it more fully trained and to have a deeper understanding of neural architectures.

\bibliographystyle{IEEEtran}
\bibliography{ref.bib}



\end{document}